\documentclass[conference]{IEEEtran}
\IEEEoverridecommandlockouts
\usepackage[T1]{fontenc}
\usepackage{mathtools}

\newcommand{\s}{{\vspace*{3mm}}}

\newcommand{\bx}{{\mathbf{x}}}
\newcommand{\by}{{\mathbf{y}}}
\newcommand{\bz}{{\mathbf{z}}}

\newcommand{\bw}{{\mathbf{w}}}

\newcommand{\bo}{{\mathbf{0}}}

\newcommand{\bs}{{\mathbf{s}}}
\newcommand{\bvf}{{\mathbf{f}}}
\newcommand{\bvh}{{\mathbf{h}}}
\newcommand{\T}{{^\top}}

\newcommand{\bR}{{\mathbf{R}}}

\newcommand{\bI}{{\mathbf{I}}}

\newcommand{\rv}[1]{\boldsymbol{#1}}

\usepackage{hyperref}
\usepackage{xcolor}
\hypersetup{
    colorlinks,
    linkcolor={black},
    linkcolor={black},
    citecolor={black},
    urlcolor={blue!80!black}
}
\usepackage{booktabs}
\usepackage{multirow}
\usepackage{cite}
\usepackage{amsmath,amssymb,amsfonts}
\usepackage{algorithmic}
\usepackage{graphicx}
\usepackage{textcomp}
\usepackage{xcolor}
\def\BibTeX{{\rm B\kern-.05em{\sc i\kern-.025em b}\kern-.08em
    T\kern-.1667em\lower.7ex\hbox{E}\kern-.125emX}}

\usepackage[caption=false]{subfig}
\usepackage{algorithmic}

\usepackage{graphicx,epstopdf}

\usepackage{subfig}
\usepackage{amsmath}
\usepackage{amssymb}
\usepackage{graphicx}
\usepackage{dcolumn}
\usepackage{mathtools}
\usepackage{amsmath,amsfonts}
\usepackage{bm}
\usepackage{amsmath}
\usepackage{amssymb}
\usepackage{color}
\usepackage{float}
\usepackage{setspace}
\usepackage{tabularx}
\usepackage{enumitem}
\AtBeginDocument{%
  \providecommand\BibTeX{{%
    Bib\TeX}}}

\begin{document}

\title{A Scalable Real-Time Data Assimilation Framework for Predicting Turbulent Atmosphere Dynamics\thanks{This manuscript has been authored by UT-Battelle, LLC, under contract DE-AC05-00OR22725 with the US Department of Energy (DOE). The US government retains and the publisher, by accepting the article for publication, acknowledges that the US government retains a nonexclusive, paid-up, irrevocable, worldwide license to publish or reproduce the published form of this manuscript, or allow others to do so, for US government purposes. DOE will provide public access to these results of federally sponsored research in accordance with the DOE Public Access Plan.}}


\author{\IEEEauthorblockN{Junqi Yin}
\IEEEauthorblockA{
\textit{National Center for Computational Sciences} \\
\textit{Oak Ridge National Laboratory}\\
Oak Ridge, TN 37831\\
yinj@ornl.gov}
\and
\IEEEauthorblockN{Siming Liang}
\IEEEauthorblockA{
\textit{Department of Mathematics} \\
\textit{Florida State University}\\
Tallahassee, FL 32306\\
sliang@fsu.edu}
\and
\IEEEauthorblockN{Siyan Liu}
\IEEEauthorblockA{
\textit{Computational Sciences and Engineering Division} \\
\textit{Oak Ridge National Laboratory}\\
Oak Ridge, TN 37831\\
lius1@ornl.gov}
\and
\IEEEauthorblockN{Feng Bao}
\IEEEauthorblockA{
\textit{Department of Mathematics} \\
\textit{Florida State University}\\
Tallahassee, FL 32306\\
fbao@fsu.edu}
\and
\IEEEauthorblockN{Hristo G.~Chipilski}
\IEEEauthorblockA{
\textit{Department of Scientific Computing} \\
\textit{Florida State University}\\
Tallahassee, FL 32306\\
hchipilski@fsu.edu}
\and
\IEEEauthorblockN{Dan Lu}
\IEEEauthorblockA{
\textit{Computational Sciences and Engineering Division}\\
\textit{Oak Ridge National Laboratory}\\
Oak Ridge, TN 37831\\
lud1@ornl.gov}
\and
\IEEEauthorblockN{Guannan Zhang}
\IEEEauthorblockA{
\textit{Computer Science and Mathematics Division} \\
\textit{Oak Ridge National Laboratory}\\
Oak Ridge, TN 37831\\
zhangg@ornl.gov}
}

\maketitle

\begin{abstract}
The weather and climate domains are undergoing a significant transformation thanks to advances in AI-based foundation models such as FourCastNet, GraphCast, ClimaX and Pangu-Weather. While these models show considerable potential, they are not ready yet for operational use in weather forecasting or climate prediction. This is due to the lack of a data assimilation method as part of their workflow to enable the assimilation of incoming Earth system observations in real time. This limitation affects their effectiveness in predicting complex atmospheric phenomena such as tropical cyclones and atmospheric rivers. To overcome these obstacles, we introduce a generic real-time data assimilation framework and demonstrate its end-to-end performance on the Frontier supercomputer. 
This framework comprises two primary modules: an ensemble score filter (EnSF), which significantly outperforms the state-of-the-art data assimilation method, namely, the Local Ensemble Transform Kalman Filter (LETKF); and a vision transformer-based surrogate capable of real-time adaptation through the integration of observational data. The ViT surrogate can represent either physics-based models or AI-based foundation models.
We demonstrate both the strong and weak scaling of our framework up to 1024 GPUs on the Exascale supercomputer, Frontier. Our results not only illustrate the framework's exceptional scalability on high-performance computing systems, 
but also demonstrate the importance of supercomputers in real-time data assimilation for weather and climate predictions.
Even though the proposed framework is tested only on a benchmark surface quasi-geostrophic (SQG) turbulence system, it has the potential to be combined with existing AI-based foundation models, making it suitable for future operational implementations. 
\end{abstract}

\section{Introduction}\label{sec:intro}
The field of meteorology is undergoing a significant transformation thanks to rapid advances in artificial intelligence (AI). For example, the European Centre for Medium-Range Weather Forecasts (ECMWF) is pioneering a new weather prediction capability referred to as the Artificial Intelligence/Integrated Forecasting System (AIFS), which was officially released in October 2023. Their approach utilizes AI emulators -- deep-learning models which predict the weather evolution by analyzing historical data that contain implicit knowledge about the governing physical laws. This enables quick and efficient forecasts on regular computers and represents a significant advantage over the demanding computations on massively parallel high-performance computing systems. While existing AI-based foundation models such as FourCastNet \cite{FourCastNet}, GraphCast \cite{graphcast}, ClimaX \cite{nguyen2023climax} and Pangu-Weather \cite{Bi2023} show considerable potential, they are not ready yet for a fully operational implementation since they are decoupled from operational data assimilation (DA) algorithms. This limitation hinders their ability to dynamically incorporate real-time observational data and impacts their effectiveness in predicting complex atmospheric phenomena, such as tropical cyclones and atmospheric rivers. The reliance on physics-based models to provide the initial conditions significantly increases the overall computational costs. In the case of AIFS, one still needs to combine the physics-based ECMWF model (IFS) with a four-dimensional DA (4D-Var) method in order to initialize the data-driven forecasts every 12h.

Data assimilation is crucial for making reliable weather forecasts because it involves the integration of real-time observational data with weather models, ensuring the models start from the most accurate representation of the current state of the Earth system. This process significantly enhances the accuracy of weather predictions by correcting discrepancies between model forecasts and real-time observations, leading to more skilful weather predictions. Moreover, DA helps in the detection and correction of model biases, improving the overall performance and reliability of weather prediction models over time. Within the Earth sciences, the ensemble Kalman filter (EnKF) of Evensen \cite{evensen_1994} and its many variants are a state-of-the-art (SOTA) DA method. Even more traditional DA systems which rely on variational algorithms use ensemble techniques to better quantify the underlying forecast uncertainty \cite{ensemble_of_DA}. EnKF methods are deployed operationally \cite{houtekamer_et_al_2014,schraff_et_al_2016} and widely used to integrate observations for the purpose of understand complex processes such as atmospheric convection \cite{aksoy_et_al_2009,aksoy_et_al_2010,jones_et_al_2016,chipilski_et_al_2020,chipilski_et_al_2022,hu_et_al_2023}. 
However, EnKFs suffer from fundamentallimitations as they make Gaussian assumptions in their update step, which leads to severe model bias in solving highly nonlinear systems. Previous studies have illustrated the detrimental effects of the resulting analysis biases in high-impact situations such as hurricane prediction \cite{poterjoy_2022}.

A viable alternative to EnKF is the particle filter (PFs) \cite{gordon_et_al_1993,vanLeeuwen_2009,vanLeeuwen_et_al_2019} -- a fully non-parametric method which converges to the correct Bayesian solution \cite{crisan_doucet_2002}. Although PFs emerged around the same time as the EnKF, their implementation to large models has been difficult in view of their curse of dimensionality (weight collapse). In practical terms, this means that PFs require prohibitively large ensemble sizes (number of particles) to retain long-term stability. While there have been significant advances in this direction \cite{todter_et_al_2016,poterjoy_et_al_2017,rojahn_et_al_2023}, the resulting PF approximations often provide marginal advantages over SOTA EnKFs used in operations.

To overcome these challenges, we propose a generic real-time DA framework and demonstrate its end-to-end performance on the Frontier supercomputer at the Oak Ridge Leadership Computing Facility (OLCF). This framework consists of two primary modules. The first module is an ensemble score filter (EnSF), originally developed in \cite{SF_2023,EnSF_2023}. The EnSF method leverages the score-based diffusion model \cite{song2021scorebased}, and has shown promising accuracy in estimating the state of a high-dimensional Lorenz-96 system with $\mathcal{O}(10^6)$ variables and highly nonlinear observations. Compared with existing diffusion models, the key ingredient is our training-free approach, which uses a Monte Carlo approximation to estimate the score function directly. This training-free procedure allows for a highly scalable formulation of the score-based filter that can be deployed at scale on supercomputers. The second primary module of our DA framework is a vision transformer (ViT)-based surrogate of the forecast model that could be either a physics-based model or an AI-based foundation model. The surrogate model is needed in our DA framework for two reasons. First, the EnSF requires the gradient of the forecast model to update the score function, and the gradient can be efficiently obtained from the surrogate model. Second, due to the complex nature of turbulence dynamics, the forecast model, especially the offline trained AI foundation models (e.g., FourCastNet), usually do not provide sufficient accuracy without incorporating observation data. Training a surrogate model using both the forward model and the observation data is an effective approach to reduce the prediction error \cite{bayesian_chao}. Nevertheless, the online training of the surrogate model requires the use of supercomputers to perform real-time DA. 

Our results demonstrate the proposed framework's exceptional scalability on high-performance computing systems, which is essential for eventual application to real weather and climate prediction problems. Even though the proposed framework is tested using the benchmark surface quasi-geostrophic turbulence (SQG) model, it has the potential to be combined with existing AI-driven weather models, making it suitable for operational use. Our contributions are listed as follows:
\begin{itemize}[leftmargin=10pt]
\itemsep0.2cm
    \item We introduce a generic real-time data assimilation framework for estimating turbulent dynamics, providing significantly more accurate predictions (Figure \ref{Linear_shocks_SQG}) than the state-of-the-art LETKF method.
    \item We showcase the remarkable strong and weak scaling capabilities of our proposed DA framework on the Frontier supercomputer, which demonstrates the necessity of supercomputers in real-time data assimilation operation.
    %
    \item We investigate the strategies for large-scale distributed training of ViTs, including compute-efficient kernel sizing on AMD MI250Xs, and memory-efficient data parallelisms for ViTs with billions of parameters.  
\end{itemize}

The rest of this paper is organized as follows. In Section \ref{sec:prob}, we introduce the physical SQG model and setup the data assimilation problem. Section \ref{sec:method} provides the details of the proposed framework, including the EnSF and the ViT-based surrogate model. The scalability experiments and results are given in Section \ref{sec:result}, while Section \ref{sec:con} summarizes the main findings and and briefly outlines our future research plans.

\section{Background}\label{sec:prob}
We first provide some background information about the data assimilation problem and the SQG model used to test the performance of the proposed DA framework described in Section \ref{sec:method}.

\subsection{Data assimilation}\label{sec:da}

Every DA algorithm requires a forecast model to describe how the physical system evolves over time, and a set of observations to reduce the growing forecast errors. In what follows next, we briefly outline the estimation-theoretic formulation of this process and point readers to standard textbooks (e.g., Jazwinski \cite{jazwinski_1970}) for a more comprehensive description.

Assume we work under the practical setting of having a discrete representation of our forecast model and let $k = 0, 1, ..., K$ denote the corresponding time index. The general evolution of the system can be written as

\begin{equation}\label{state}
\text{\bf Forecast model:}\quad \rv X_k = \bvf_{k-1}(\rv X_{k-1}, \rv E^m_{k-1}),
\end{equation}
\noindent where $\rv X_k$ is the discretized state. Note that this forecast model could be either physics-based like the SQG, or an AI-based foundation model like FourCastNet. We further assume the model predictions are not perfect, and their errors captured by the random vector $\rv E^m_k$. 

To correct the model predictions, we use a sequence of observations given by

\begin{equation}\label{observations}
\text{\bf Observation model:}\;\; \rv Y_{k} = \bvh_k(\rv X_k) + \rv E^o_k,
\end{equation}
where $\bvh_k$ is the observation operator mapping the state to observation space and $\rv E^o_k \sim \mathcal{N}(\bo,\bR_k)$ is the corresponding observation error. In this case, we have made the simplifying assumption that observations are additive and Gaussian in nature, but more flexible models can be also used \cite{chipilski_2023}. 

Given the forecast and observation models, a standard way to solve the DA problem is to calculate the filtering probability density function (PDF) $P(\bx_k|\by_{1:k})$, in which the state is conditioned on the entire history of observations up to the present (filtering) time. This can be done by iterating through one prediction and one update (analysis) step, as described below.

\vspace{0.1cm}
\paragraph*{\bf Prediction}
Due to its stochastic nature, the state is evolved forward using the Chapman-Kolmogorov equation such that
\begin{equation}\label{Kolmogorov}
P(\bx_k | \by_{1:k-1}) = \int P(\bx_{k-1} | \by_{1:k-1}) P(\bx_k | \bx_{k-1}) d\rv \bx_{k-1},
\end{equation}
where $P(\bx_k | \bx_{k-1})$ is the transition PDF to be determined from the forecast model ~\eqref{state}.

\vspace{0.2cm}
\paragraph*{\bf Update} 

After the new measurements $\rv Y_k = \by_k$ are collected, the error-prone forecasts are adjusted using a form of Bayes' theorem in which 

\begin{equation}\label{Bayesian}
P(\bx_k | \by_{1:k}) \propto P(\bx_k | \by_{1:k-1}) P(\by_k|\bx_k),
\end{equation}
Accounting for the additive-Gaussian assumption on the observation errors $\rv E^o_k$, the likelihood $P(\by_k|\bx_k)$ can be rewritten as
\begin{equation}\label{Likelihood}
\begin{aligned}
    P(\by_k|\bx_k) \propto 
     \exp\Big[ - (\by_k- \bvh(\bx_k))\T \bR_k^{-1} (\by_k- \bvh(\bx_k)) \Big].
\end{aligned}
\end{equation}

\subsection{The surface quasi-geostrophic (SQG) model}\label{sec:SQG}

The new prediction framework is tested on a benchmark model simulating the surface quasi-geostrophic (SQG) dynamics \cite{tulloch_smith_2009a}. The numerical implementation follows \cite{tulloch_smith_2009b} closely: it represents a nonlinear Eady model with an f-plane approximation as well as uniform stratification and shear. The spatial discretization is done in spectral space and is based on the fast Fourier transform (FFT). The time integrator uses a 4$^{\text{th}}$-order Runge Kutta scheme with a $2/3$ dealiasing rule and implicit treatment of hyperdiffusion. For more details, readers are directed to the open-source GitHub repository of the model, which can be accessed via \url{https://github.com/jswhit/sqgturb}.

It is important to emphasize that the proposed DA framework can be combined with any forecasting model, either physics-based or AI-driven, as described in Section \ref{sec:method}. Nevertheless, our choice to work with the SQG model for this study is motivated by its ability to generate turbulence behavior that is representative of real geophysical flows. In particular, fully developed turbulence in the SQG system follows a kinetic energy (KE) density spectrum with a -5/3 slope, which aligns with reference measurements from field campaigns \cite{nastrom_gage_1985}. Previous studies have shown that such turbulence characteristics set a limit on the ability to make reliable weather predictions \cite{durran_gingrich_2014}. Following the seminal work of Edward Lorenz \cite{lorenz_1969}, we know that 3D flows with this turbulence spectrum are very sensitive to errors in the initial conditions (ICs). The rapid amplification of IC uncertainty represents a barrier for how far in advance we can predict chaotic weather patterns ($\sim$2 weeks). While the SQG model is much simpler compared to operational NWP systems based on the full set of governing equations, its ability to generate realistic turbulence behavior makes it a suitable candidate for the numerical tests presented here. Crucially, our results highlight the importance of coupling AI-based forecasting methods with advanced DA techniques in order to control the errors arising in chaotic dynamical systems.

\section{Methodology}\label{sec:method}
\begin{figure*}[h!]
    \centering
    \includegraphics[width=0.8\textwidth]{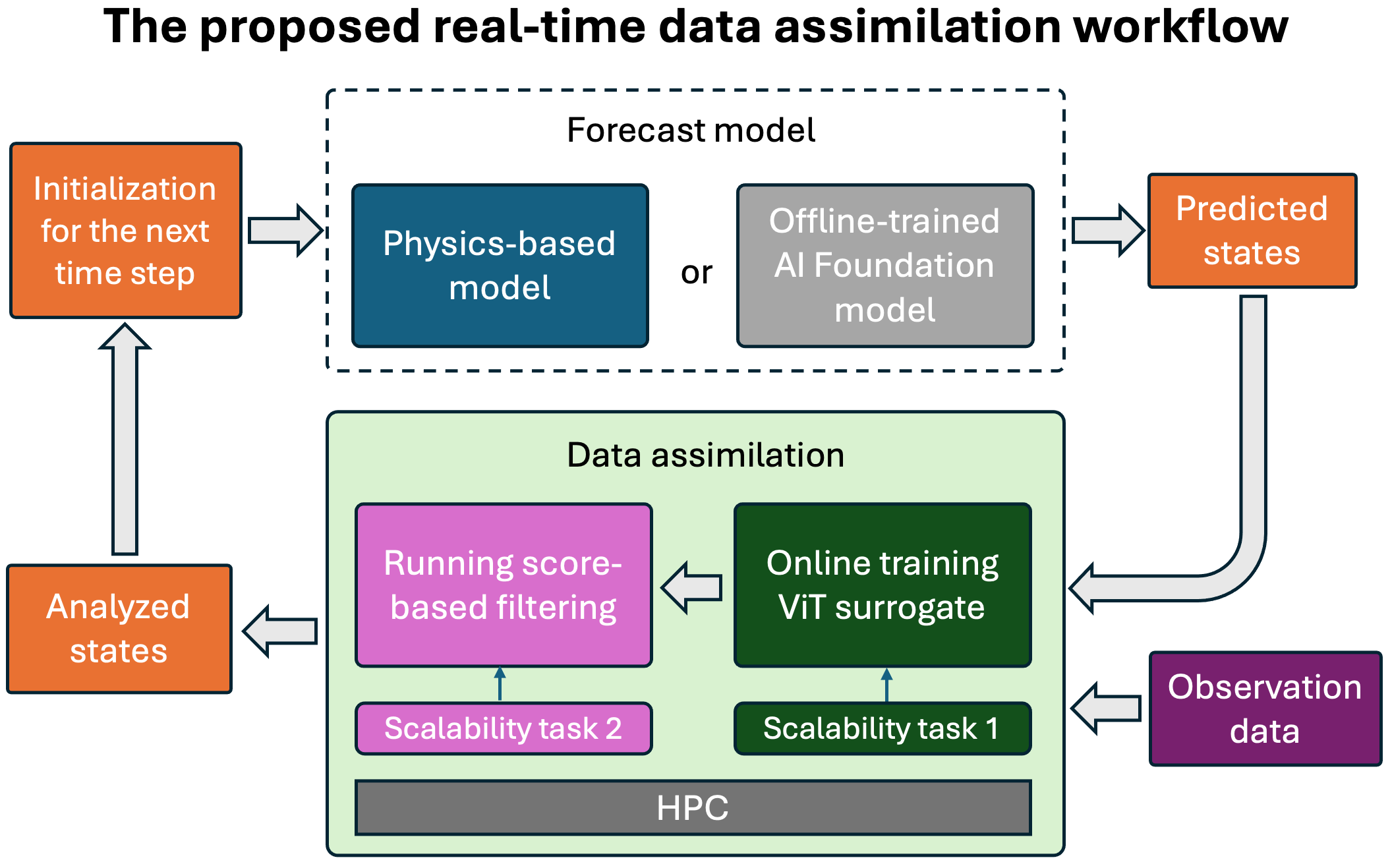}
    \vspace{-0.cm}
    \caption{Illustration of the real-time sequential DA workflow, which needs to be performed very frequently (e.g., every hour) in weather forecast operation. 
    Recent advances in weather and climate modeling focus on developing AI-based foundation models, e.g., FourCastNet, GraphCast, etc., to replace the traditional physics-based forecast models. These data-driven architectures are not yet ready for operational use due to the lack of real-time data assimilation capabilities. The proposed DA framework has two primary modules that need to be scaled on HPC, i.e., the ensemble score filter (EnSF) introduced in Section \ref{sec:filter}, which significantly outperforms SOTA methods like LETKF, and a vision transformer(ViT)-based surrogate, introduced in Section \ref{sec:ViT}, capable of real-time adaptation through the integration of observational data. Our method can be integrated with either physics models or AI-based foundation models.  
    The scalability of our method on HPC is essential to ensure computations can be performed in real time.}
    \label{fig:workflow}
\end{figure*}

This section contains the details of the proposed real-time DA framework. The corresponding workflow is summarized in Figure \ref{fig:workflow}. There are two major scalability tasks, one is the online training of the ViT surrogate using observational data, and the other is the efficient running of the EnSF. Since training ViT and running EnSF occurs sequentially with each filtering iteration, the overall computing time is the summation of the computing times for these two steps. We will describe the EnSF method in Section \ref{sec:filter} and the online training of the ViT surrogate in Section \ref{sec:ViT}, respectively.

\subsection{The ensemble score filter (EnSF)}\label{sec:filter}

The major challenge of DA for operational use is that there is no existing method that can simultaneously resolve the following three issues: nonlinearity/non-Gaussianity, high-dimensionality, and scalability on HPC. The Local Ensemble Transform Kalman Filter (LETKF) of Hunt et al.~\cite{hunt_et_al_2007} is widely used in the geophysical community because of its good scalability on HPCs. For instance, LETKF is the choice for an operational DA method in the German weather prediction system KENDA \cite{houtekamer_et_al_2014,schraff_et_al_2016}. However, it cannot effectively handle highly nonlinear/non-Gaussian DA problems like hurricane prediction. As discussed earlier, PFs can tackle arbitrarily complex problems, but they suffer from the curse of dimensionality, which makes their operational implementation quite challenging. The new DA method described next has demonstrated its ability to resolve all three issues, and has the potential to significantly improve SOTA weather and climate predictions.

\s \subsubsection{Overview of diffusion models}
To describe score-based diffusion models, we need to introduce the following stochastic differential equation (SDE)
\begin{equation}\label{forward:SDE}
d \rv{Z}_t = b(t)\rv{Z}_t dt + \sigma(t)d\rv{W}_t, 
\end{equation}
with $\rv{W}_t$ being the standard Brownian motion, whereas $b$ and $\sigma$ are the pre-defined drift and diffusion coefficients. The initial condition of the SDE $\rv{Z}_0$ follows some target distribution, which in our case is set to the filtering PDF given by Eq.~\eqref{Bayesian}. Assuming all distributions are differentiable, we will denote the PDF of this target by $Q(\bz_0)$. With properly chosen $b$ and $\sigma$, it is possible to use the diffusion process $\{\rv Z_t\}_{0\leq t \leq T}$ over the pseudo-time interval $[0, T]$ and transform any ${Q}(\bz_0)$ to the standard Gaussian distribution $\mathcal{N}(\bo, \bI)$. In particular, the following reverse-time SDE can be used to generate samples $\{\rv Z_0^i \}_{i=1}^{N}$ of the target random vector $\rv Z_0$:
\begin{equation}\label{reverse:SDE}
\begin{aligned}
d\rv Z_t = \left[ b(t)\rv Z_t -\sigma^2(t) \bs(\rv Z_t,t) \right] dt &+ \sigma(t)d\overleftarrow{\rv W}_t 
\end{aligned}
\end{equation}
where we have used the notation $\int \cdot d\overleftarrow{\rv W}_t$ to define a {backward} It\^o stochastic integral \cite{SDE, BDSDE}. Within this new SDE, the term $\bs(\cdot, t)$ is referred to as the {score function} and is a short-hand for
\begin{equation}\label{eq:score1}
    \bs(\bz_t, t) = \nabla \log(Q(\bz_t)).
\end{equation}

It is worth mentioning that the score function $\bs(\cdot, t)$ is an essential ingredient for transforming the standard Gaussian distribution of $\rv Z_T$ to the target distribution $Q(\bz_0)$. Furthermore, once the score function corresponding to the target PDF $Q(\bz_0)$ is obtained, we can generate an unlimited number of Gaussian samples (a computationally efficient process) and use them as an input to the reverse-time SDE in ~\eqref{reverse:SDE} to get an unlimited number of samples from the complex target distribution. One important technicality is that the drift and diffusion coefficients $b$ and $\sigma$ need to be properly chosen in order to obtain the desired transformation. Here we follow \cite{song2021scorebased} and define these functions as
\begin{equation}\label{coefficients}
\begin{aligned}
    &b(t) = \frac{d\log \alpha_{t}}{dt},
&\sigma^2(t) = \frac{d \beta^2_{t}}{d t} - 2 \frac{d \log \alpha_{t}}{d t} \beta^2_{t},
\end{aligned}
\end{equation}
with $\alpha_{t} = 1 - t$ and $\beta_{t} = \sqrt{t}$ for $t \in [0, 1]$.

\s
\subsubsection{The ensemble score filter (EnSF)}\label{sec:ensf}
The main philosophy behind EnSF, our new filtering approach, is to approximate the score functions $\bs_{k | k-1}$ and $\bs_{k | k}$ corresponding to the prior (forecast) and posterior PDFs in \eqref{Kolmogorov} and \eqref{Bayesian}. Let us first suppose we have access to the posterior score $\bs_{k-1|k-1}$ at the previous time level $k-1$. After generating a standard Gaussian sample $\{\rv Z^m_{t_{N}}\}_{m=1}^M \sim \mathcal{N}(\bo, \bI)$, we can pass each sample through an appropriately discretized version of the reverse-time SDE in \eqref{reverse:SDE} (e.g., an Euler scheme) to produce the analysis ensemble $\{\rv X^{m}_{k-1|k-1}\}_{m=1}^{M}$ from the desired Bayesian posterior $P(\bx_{k-1} | \by_{1:k-1})$. Since Gaussian sampling is a computationally efficient process, we can get a large number of target samples for a more adequate description of the posterior uncertainty. In general, the choice of the ensemble size $M$ will be determined by the complexity of the specific application or the available computational resources.

Once we obtain the analysis ensemble at time level $k-1$, the EnSF's workflow reduces to the standard iterative application of prediction and update steps, as described next.

\paragraph*{\bf Prediction step} This part of the algorithm is identical for all ensemble-based approaches and uses the forecast model \eqref{state} on each analysis member $\rv X^{m}_{k-1|k-1}$, with the integration length determined by the time separation between observations. The resulting sample $\{\rv X^{m}_{k|k-1} \}_{m=1}^M$ represents an unbiased approximation of the prior PDF $P(\bx_k | \by_{1:k-1})$ and will be utilized in the estimation of the prior score $\hat{\bs}_{k| k-1}$.

\paragraph*{\bf Update step} The main goal here is to obtain an approximation for the posterior score $\bs_{k | k}$ (i.e., $\hat{\bs}_{k| k}$). Using \eqref{Bayesian}, we first recognize that the Bayesian posterior is proportional to the prior-likelihood product. Relating this expression to score functions simply requires us to take the gradient of the logarithm of \eqref{Bayesian}:
\begin{equation}
\begin{aligned}
 &\nabla_\bx \log P(\bx_{k} | \by_{1:k}) \\
 = &\nabla_\bx \log P(\bx_k | \by_{1:k-1}) +  \nabla_\bx \log P(\by_k | \bx_k),
\end{aligned}
\end{equation}
In the above expression, we identify $\nabla_\bx \log P(\bx_k | \by_{1:k})$ as the posterior score function $\bs_{k | k}(\bz, t)$ to be estimated, whereas $\bs_{k|k-1}(\bz, t) \coloneqq \nabla_\bx \log P(\bx_k | \by_{1:k-1})$ is the prior score calculated during the prediction step. Analogously, the last term represents the likelihood score and determines how observations should be incorporated during the update step. In EnSF, we implement a slightly modified version of the posterior score $\bs_{k | k}(\bz, t)$ such that 
\begin{equation}\label{S-posterior}
\bs_{k | k}(\bz, t) := \bs_{k|k-1}(\bz, t) + h(t) \nabla_\bx \log p(\by_k | \bz).
\end{equation}
Note that the coefficient $h(t)$ multiplying the likelihood score represents a damping factor that decreases over the pseudo-time interval $[0,T]$ such that $h(0) = 1$ and $h(T) = 0$. In our numerical experiments, we define $h(t) = T - t$, although other options are also possible and will be explored in future work.

Since the likelihood score can be calculated analytically due to the Gaussian assumptions in \eqref{Likelihood}, $\bs_{k | k}(\bz, t)$ is readily obtained as soon as we finish estimating $\bs_{k|k-1}(\bz, t)$ from the forecast ensemble $\{\rv X^{m}_{k|k-1} \}_{m=1}^M$. As explained earlier, this is accomplished with a training-free procedure that replaces the standard deep learning techniques used for estimating scores in diffusion models \cite{song2021scorebased, SF_2023}. The starting point is to set the target random $\rv Z_0$ be the forecast ensemble $\{\rv X^{m}_{k-1|k-1}\}_{m=1}^{M}$. Using the score function definition and leveraging the forms of the drift and diffusion coefficients $b$ and $\sigma$, the conditional PDF $Q(\bz_t|\bz_0)$ needed in the forward SDE \eqref{forward:SDE} can be written as

\begin{equation}
Q(\bz_t|\bz_0) \propto \exp\Big[ - \frac{1}{2 \beta_t^2}  (\bz_t - \alpha_t \bz_0)\T (\bz_t - \alpha_t \bz_0) \Big].
\end{equation}
Marginalizing over $\bz_0$ gives the following score function
\begin{equation}\label{eq:score}
\begin{aligned}
& \bs(\bz_{t}, t)\\
  =& \nabla_\bz \log Q(\bz_t) = \nabla_\bz \log \left(\int Q({\bz}_t | \bz_0) Q(
\bz_0) d\bz_0\right)\\[2mm]
 = &\frac{1}{\int Q(\bz_t | \bz'_0) Q(
\bz'_0) d\bz'_0}  \int  - \frac{\bz_t - \alpha_t \bz_0}{\beta^2_t} Q(\bz_t | \bz_0) Q(\bz_0) d\bz_0\\[2mm]
 =&  - \int \frac{\bz_t- \alpha_t \bz_0}{\beta^2_t} \bw_t(\bz_t,  \bz_0)  Q(\bz_0)d\bz_0.
\end{aligned}
\end{equation}
Notice that the weight function $\bw_t(\bz_t,  \bz_0)$ follows the definition
\begin{equation}\label{eq:weight}
\bw_t(\bz_t,\bz_0) =  \frac{ Q(\bz_t | \bz_0) }{\int Q(\bz_t | \bz'_0) Q(\bz'_0) d\bz'_0},
\end{equation}
and satisfies the condition $\int \bw_t(\bz_t,  \bz_0) Q(\bz_0) d\bz_0 = 1$. 

Finally, we utilize the form of \eqref{eq:score} to perform a Monte Carlo approximation of $\bs_{k | k-1}$ for a given $\bz$ and $t \in [0, 1]$ such that
\begin{equation}\label{Approx:S-prior}
\begin{aligned}
    \bs_{k|k-1}(\bz, t) \approx &\; \hat{\bs}_{k|k-1}(\bz, t) \\
    = & \sum_{j=1}^{J} - \frac{\bz - \alpha_{t} \bx_{k|k-1}^{m_j}}{\beta^2_{t}} \hat{\bw}_{t}\left(\bz, \bx_{k|k-1}^{m_j} \right),
\end{aligned}
\end{equation}
where $\{\rv X_{k|k-1}^{m_j}\}_{j=1}^J$ represents a mini-batch from the forecast ensemble $\{\rv X_{k|k-1}^m\}_{m=1}^M$. On the other hand, $\bar{\bw}_{t}$ is another Monte Carlo approximation of the weight $\bw_{t}$ computed from
\begin{equation}\label{weight_app}
    \hat{\bw}_{t}\left(\bz, \bx_{k|k-1}^{m'_j}\right)  = \frac{Q\left(\bz | \bz_{k|k-1}^{m'_j} \right)}{\sum_{j=1}^J Q\left(\bz | \bx_{k|k-1}^{m_j} \right)}.
\end{equation}

After we solve for \eqref{Approx:S-prior}, it is straightforward to obtain the approximate posterior score from \eqref{S-posterior}:  
\begin{equation}\label{Approx:S-posterior}
\hat{\bs}_{k | k}(\bz, t) = \hat{\bs}_{k|k-1}(\bz, t) + h(t) \nabla_\bx \log P(\by_k | \bz).
\end{equation}
Completing the update step then pertains to running the discretized reverse-time SDE with $\hat{\bs}_{k | k}$ and storing the output as the desired analysis ensemble $\{\rv X^{m}_{k|k}\}_{m=1}^{M}$.

\s \subsubsection{Scalable implementation of EnSF on HPC}\label{sec:ensf_hpc}
We have implemented the EnSF method in PyTorch, making the code base compatible with both CPU-based platforms and those equipped with accelerators. The computational workload scales with various factors, including the number of ensembles, problem dimensions, and the total number of filtering cycles. The most efficient factor for parallelization are the ensembles, as it incurs minimal communication overhead. Considering the large memory capacity of GPUs on Frontier, straightforward parallelization can already support EnSF with dimensions up to 100 million, which is more than sufficient for our application. Since the training of the ViT surrogate is the bottleneck of the overall scaling, we will focus on the optimization of distributed training in the following.         

\subsection{ViT surrogate for the SQG model}\label{sec:ViT}
\begin{figure}[h!] 
\begin{center}
\includegraphics[width = 0.48\textwidth]{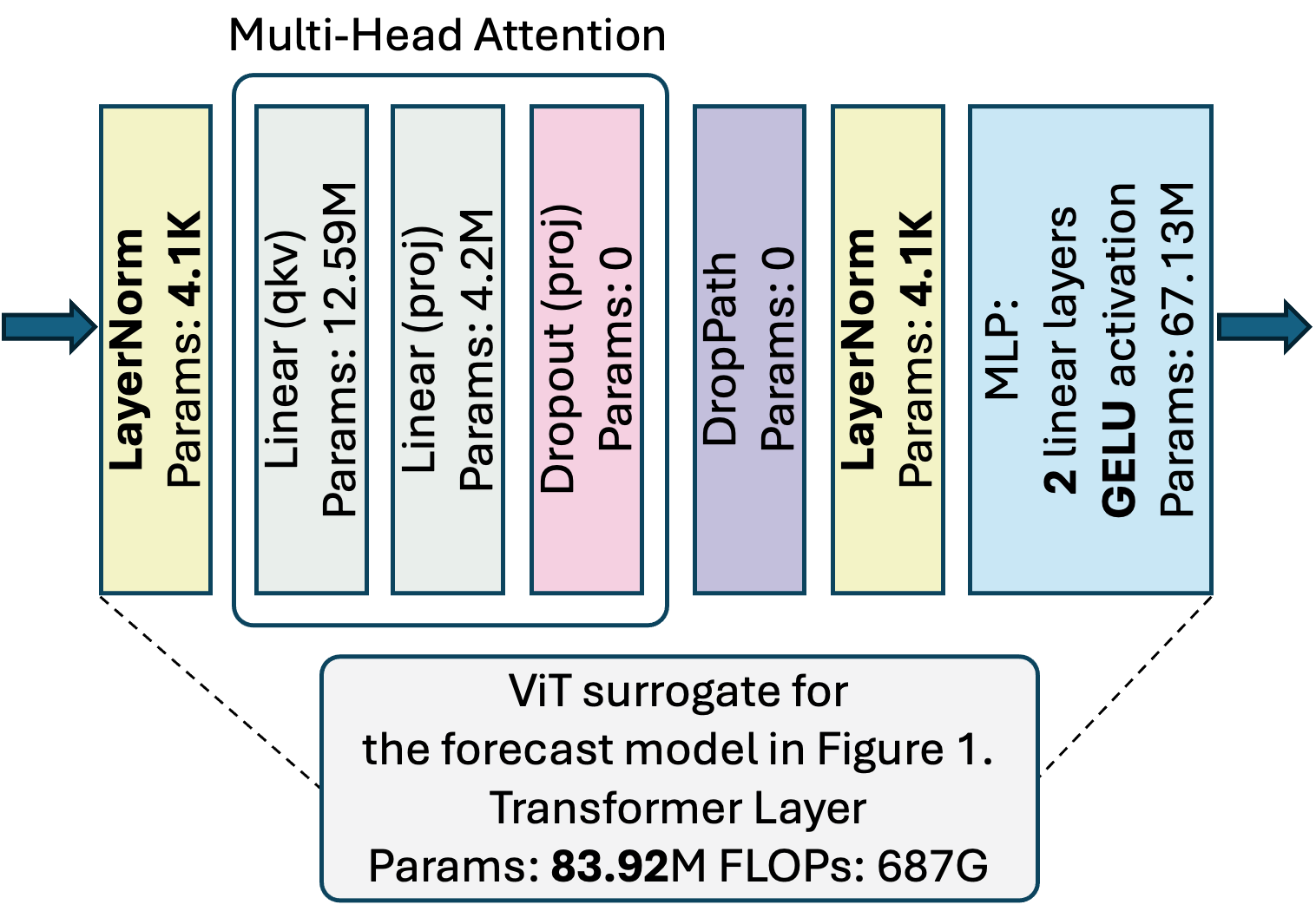}
\end{center}
\caption{Building block of ViT surrogate model for the forecast model in Figure \ref{fig:workflow}. The number of parameters and floating point operations (FLOPs) are exemplified with 8-head attention, an embedding dimension of 2048, and a MLP to attention ratio of 8.}\label{fig:arch}
\end{figure}

\paragraph{\bf Compute-efficient architecture}
We have developed a Vision Transformer (ViT) surrogate tailored specifically for the surface quasi-geostrophic (SQG) model, utilizing a standard ViT backbone. Figure~\ref{fig:arch} illustrates the architecture of SQG-ViT, which consists of multi-head self-attention and multi-layer perceptron (MLP) components, augmented by normalization layers before and after the attention mechanism. To address overfitting, we have incorporated Dropout and DropPath regularization techniques. It is worth noting that the MLP component typically dominates the parameter count, making matrix-matrix multiplication (GEMM) the most computationally intensive operation.

The performance of GEMM is significantly influenced by the shapes of the matrices \cite{spock,co-design}, thereby impacting the overall training efficiency of ViT. This dependency underscores the importance of appropriately sizing kernels, a task determined by factors such as embedding dimension, number of attention heads, and the ratio of MLP to attention. Adhering to the scaling law for Transformer architecture, where model capacity scales with the number of parameters, optimizing kernel sizes for computational efficiency becomes imperative for large-scale training on high-performance computing (HPC) systems. Such optimization not only reduces computational load but also conserves energy, promoting sustainable computing practices. In the following section, we describe our distributed training strategies.
%
\begin{table}[h!]
\small
\begin{tabular}{c|cccc}
\toprule
Method & optimizer        & \begin{tabular}[c]{@{}c@{}}optimizer\\ gradient\end{tabular} & \begin{tabular}[c]{@{}c@{}}optimizer\\ gradient\\ weight\end{tabular} & hierarchical  \\ \midrule
FSDP   & n/a  & shard\_grad\_op                                                         & full\_shard                                                           & hybrid\_shard \\
ZeRO   & stage 1         & stage 2                                                      & stage 3                                                               & n/a           \\ \bottomrule
\end{tabular}
\caption{The distributed training methods with different memory partition strategies.} \label{tab:dist}
\vspace{-0.3cm}
\end{table}

\paragraph{\bf Fully sharded data parallel (FSDP)}
In addition to conventional data parallelism, where each device hosts a duplicate of the model, recent advancements in memory-efficient data parallelism, such as FSDP, have emerged as more suitable options for training large models due to their reduced memory footprint. Even when utilizing half precision, Vision Transformer (ViT) training necessitates approximately 12 times the model parameter size in memory storage, encompassing model weights (1X), optimizer states (2X for Adam optimizer), gradients (1X), and intermediate storage (2X) like FSDP units. FSDP offers distributed partitioning of various memory components through three strategies outlined in Table~\ref{tab:dist}. Specifically, \texttt{shard\_grad\_op} distributes gradients and optimizer states across all devices, \texttt{full\_shard} partitions all memory components, and \texttt{hybrid\_shard} represents a blend of data parallelism and FSDP. Due to the \texttt{AllGather} operation for partitions, FSDP incurs approximately 50\% more communication volume compared to data parallelism, although some of this overhead can be absorbed by computational operations.

\paragraph{\bf ZeRO data parallel}
Besides PyTorch built-in FSDP, another widely utilized memory-efficient data parallel implementation is DeepSpeed ZeRO. These two strategies exhibit an almost one-to-one correspondence (refer to Table~\ref{tab:dist}). However, ZeRO offers a broader array of tuning parameters for performance optimization compared to FSDP. These include adjusting the message bucket size for operations like \texttt{AllGather} and \texttt{Reduce}, enabling continuous memory allocation for gradients, and other similar optimizations.

\paragraph{\bf Computational budget estimation}
The total number, T, of floating-point operations (FLOPs) required for training ViT is directly proportional to the number of tokens, which depends on factors such as input size (L), patch size (P), number of epochs (E), and the number of model parameters (M). Specifically, this relationship follows
\begin{equation}
    T = 6 \prod_{i=1}^d\frac{L_i}{P_i}*E * M,
\end{equation}
where $d$ represents the dimension of the input image. The number of tokens per input image is given by the product, and hence $T$ is essentially proportional to the total number of tokens during the training and the number of model parameters. The factor 6 comes from the fact that every token is processed with a multiply-accumulate (MAC) and two MACs during the forward and backward propagation, respectively.  
\begin{figure}[h!] 
\begin{center}
\includegraphics[width=0.48\textwidth]{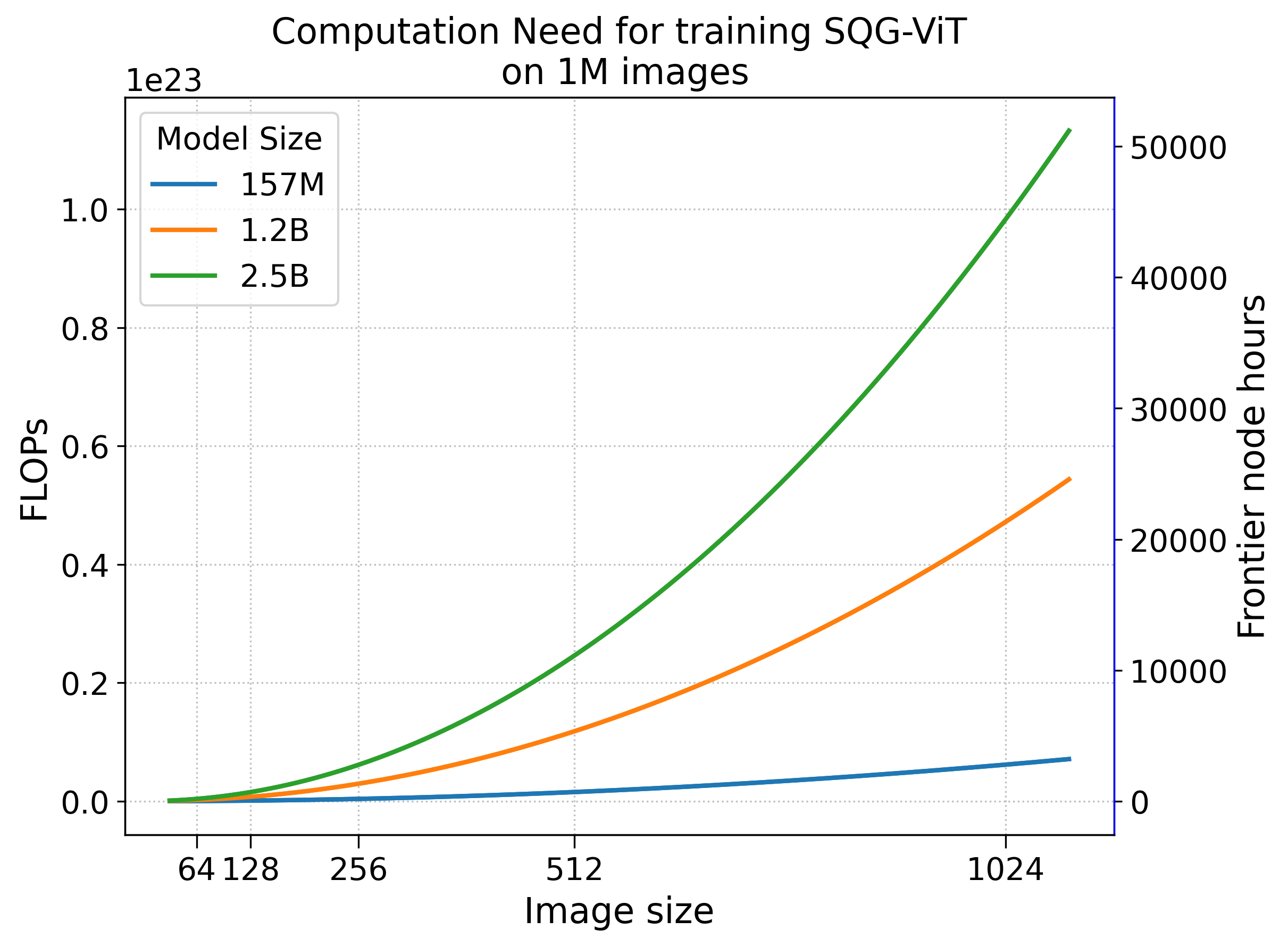}
\end{center}
\vspace{-0.75em}
\caption{Computation need in terms of FLOPs and Frontier node hours for training ViT surrogate model for the SQG model on 1M images. }\label{fig:cost}
\end{figure}
In Figure~\ref{fig:cost}, we present the total number of FLOPs and the computation hours (in the unit of Frontier node hours) needed to train three representative sizes of ViT. Without loss of generalizability, we assume training over 100 epochs with a dataset containing 1 million images.  

\section{Results}\label{sec:result}
We perform the experiments on the first Exascale supercomputer, Frontier. Each Frontier node is equipped with four AMD Instinct MI250X GPUs with dual Graphics Compute Dies (GCDs) and one third-generation EPYC CPU. A GCD is viewed as an effective GPU, and we use GCD and GPU interchangeably in the following discussion. All four MI250Xs (eight effective GPUs) are connected using 100 GB/s Infinity Fabric (200 GB/s between 2 GCDs of MI250X), and the nodes are connected via a Slingshot-11 interconnect with 100 GB/s of bandwidth. Frontier consists of 9408 nodes in total, i.e., 75,264 effective GPUs (each equipped with 64GB high-bandwidth memory). 
We report the following two sets of experimental results:
\begin{itemize}[leftmargin=15pt]
\itemsep0.2cm
    \item {\bf Accuracy tests}: Comparing our method with the state-of-the-art LETKF method to demonstrate the superior accuracy of our method in predicting highly nonlinear turbulent dynamics.
    \item {\bf Scalability tests}: Demonstrating the scalability of the proposed real-time DA workflow in Figure \ref{fig:workflow}, including the online ViT training and the online EnSF execution. 
\end{itemize}


\subsection{Accuracy tests}\label{sec:accuracy}
\paragraph{\bf The state-of-the-art DA method for comparison} 
LETKF is a deterministic (square-root) EnKF method which was originally proposed by Bishop et al.~\cite{bishop_et_al_2001} and further developed in Hunt et al.~\cite{hunt_et_al_2007}. The reason why these algorithms are preferred at operational scales is their embarrassingly parallel structure. In particular, the LETKF update equations can be applied independently within local regions surrounding individual grid points. The size of each region is typically determined through the cut-off radius in correlation functions (e.g., Gaspari-Cohn \cite{gaspari_cohn_1999}). For our SQG implementation, the horizontal and vertical extents of each local domain are dynamically coupled through the Rossby radius of deformation \cite{wang_et_al_2021}. Additional regularization strategies include the distance-dependent inflation of observation errors (R-localization) as well as the relaxation to prior spread (RTPS) inflation \cite{whitaker_hamill_2012}. As usual, the localization (cut-off) radius and inflation factors are optimally tuned to minimize the LETKF's analysis errors.

\paragraph{\bf Experimental setup} 
For our numerical tests, we discretize the SQG model on a 64x64x2 mesh and evaluate the errors of different DA systems in the setting where the entire SQG state is directly observed; that is, the observation operation $\bvh_k$ in \eqref{observations} becomes the identity matrix $\bI$. For simplicity, the error covariance matrix $\bR$ is also set to $\bI$. Observations are generated synthetically every 12h within a standard observation system simulation (OSSE) framework \cite{hoffman_atlas_2016}. 

We also consider the imperfect model scenario in which we add random model errors drawn from an uncorrelated Gaussian distribution (i.e., diagonal covariance matrix). The errors white in time, but are comprised of four stochastic processes characterized by a different probability of occurrence and amplitude -- $20\%, 15\%, 10\%$ and $5\%$ chance of realization with amplitudes equal to $20\%, 30\%, 40\%$ and $50\%$ of the average SQG model values, respectively. The purpose of introducing external model errors is twofold: (i) to create a more challenging testbed for our new data assimilation framework, and (ii) reflect the typical scenario in which real weather and climate models are subject to unexpected errors due to their simplified formulation. 

The ensemble size for both DA algorithms (LETKF and EnSF) is set to 20. Initial ensembles are created through the random selection of model states from a long-term integration of the SQG model. Since the external model errors discussed earlier are unpredictable, LETKF's inflation and localization parameters are tuned in an error-free twin experiment. We find that the optimal RTPS factor and cut-off localization scales are 0.3 and 2000 km, respectively. One significant advantage of the EnSF algorithm used in our new DA framework is that it can maintain stable performance without any special tuning. For the numerical tests presented in this study, localization is not applied and the variance (spread) of the analysis ensemble is simply relaxed to the prior (forecast) values in order to guarantee the long-term filter stability.

We compare the performance over the time period $t  \in [0,3600]$ and consider the four different architectures:
\vspace{0.1cm}
\begin{itemize}[leftmargin=20pt]
\itemsep0.2cm
    \item {\bf SQG only}: Run the SQG model iteratively from $t = 0$ to $t = 3600$ without incorporating observations. 
    \item {\bf ViT only}: Run the offline trained ViT surrogate iteratively from $t = 0$ to $t = 3600$ without using observations. 
    \item {\bf SQG + LETKF}: Apply LETKF (a SOTA method in the DA community) to assimilate observations and correct the SQG forecasts.
    %
    \item {\bf ViT + EnSF}: This is the proposed framework in this study -- use the more accurate EnSF method to adjust the forecasts from the pre-trained ViT surrogate of the true SQG dynamics.
\end{itemize}

\begin{figure}[h!]
\centering
\includegraphics[width = 0.48\textwidth]{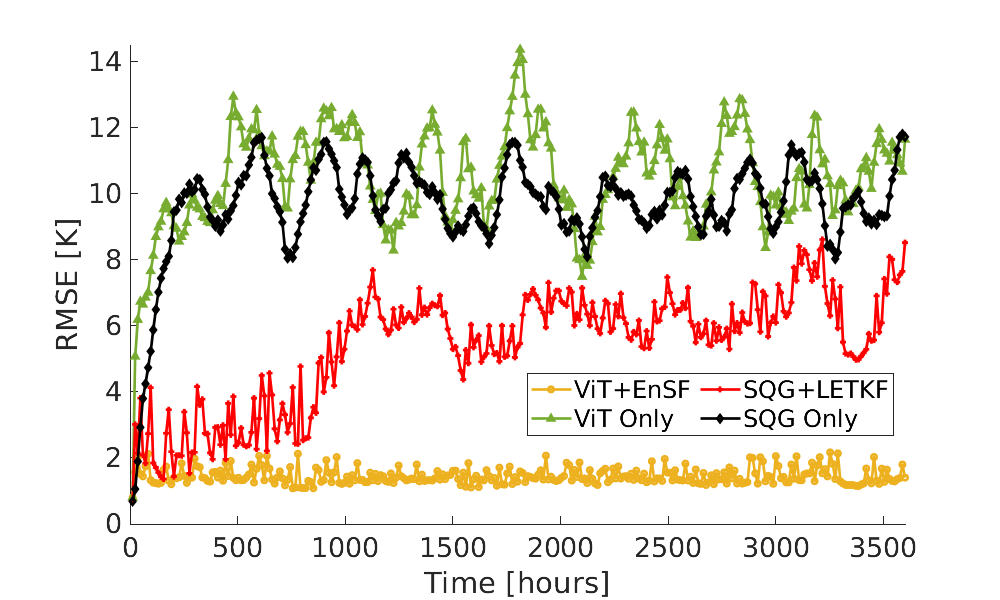}
\vspace{-0.2cm}
\caption{The root mean squared errors (RMSEs) of the four test cases. We observe that data assimilation is a necessary component to ensure accurate reconstruction of the SQG state. On the other hand, the RMSE of experiments that only use SQG or ViT without a DA component grows very fast in time. Moreover, LETKF diverges from the ground truth as model errors accumulate in time, suggesting that the LETKF method is sensitive to model imperfections. The proposed EnSF+ViT framework provides superior performance since we observe stable performance throughout all analysis cycles even in the absence of fine tuning.}
\label{Linear_shocks_SQG}
\end{figure}

\begin{figure*}[t!]
\includegraphics[width=0.98\textwidth]{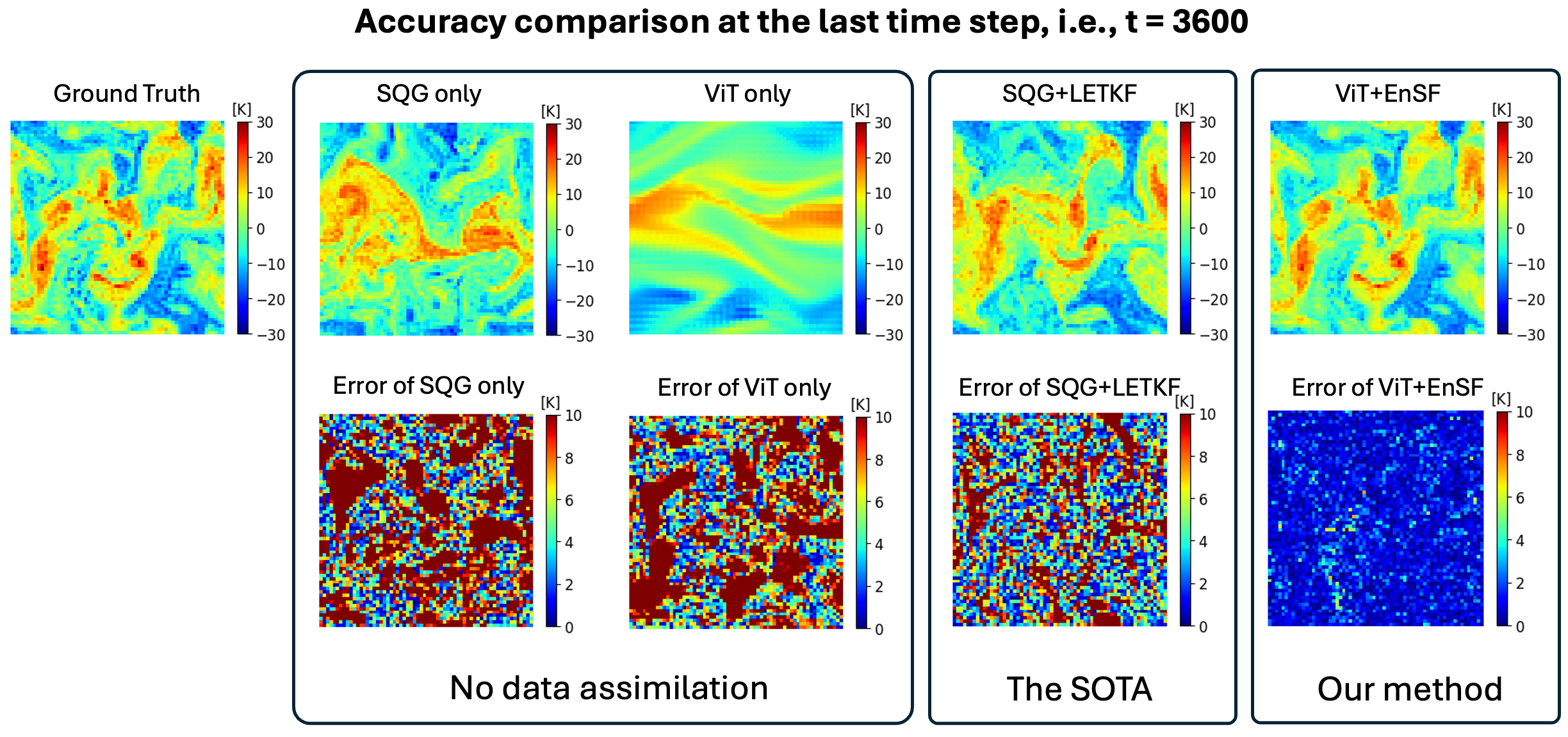}
\caption{The top row shows the analysis ensemble means from SQG only, ViT only, LETKF+SQG and EnSF+ViT with respect to the ground truth potential temperature field at the final observation time, i.e., $t = 3600$. The analysis mean errors of the four experiments are displayed on the bottom row. We confirm that pure physics-based or AI-based model predictions without data assimilation cannot provide an accurate long-term state reconstruction of the SQG state due to the rapid growth of initial errors in chaotic dynamical systems. The SOTA LETKF method captures the overall large-scale pattern but fails to represent small-scale features. The proposed EnSF+ViT offers the best accuracy, consistent with the RMSE statistics shown in Figure \ref{Linear_shocks_SQG}.}
\label{Linear_2D_Error}
\end{figure*}
Figure \ref{Linear_shocks_SQG} shows the root mean squared error (RMSE) of the above four experiments. We can make several important observations. First, DA is a necessary component to ensure accurate long-term reconstruction of the SQG state. This to be contrasted with the SQG-only and ViT-only experiments where the RMSEs experience a rapid growth as a result of the developing SQG turbulence. This is caused by the chaotic dynamics and the rapid amplification of IC errors. Second, the LETKF RMSEs gradually increase as we add model errors to true SQG state. Eventually, the LETKF's performance is comparable to the SQG-only and ViT-only simulations in which DA is not carried out. The latter implies that the SOTA LETKF method is sensitive to model imperfections even when the inflation and localization parameters are optimally tuned. Third, EnSF+ViT provides superior performance -- we observe stable results throughout the entire integration period without any special fine tuning.  

To visualize differences between the four methods, Figure \ref{Linear_2D_Error}
displays snapshots of the analysis ensemble means and the corresponding errors during the last integration time, $t = 3600$. The top row illustrates that the proposed EnSF+ViT method (last column) is much closer to the ground truth . While the SOTA LETKF+SQG manages to capture the large-scale eddy features, it cannot adequately represent some of their their fine-scale details (e.g., the extreme temperature values).

\subsection{Scalability tests} 
We investigate the scaling of the proposed DA framework, i.e., the ViT+EnSF workflow, on Frontier from the compute-efficient architecture search on single node, to performance analysis and profiling, and optimization at scale.  

\begin{figure}[h!] 
\centering
\includegraphics[width = 0.48\textwidth]{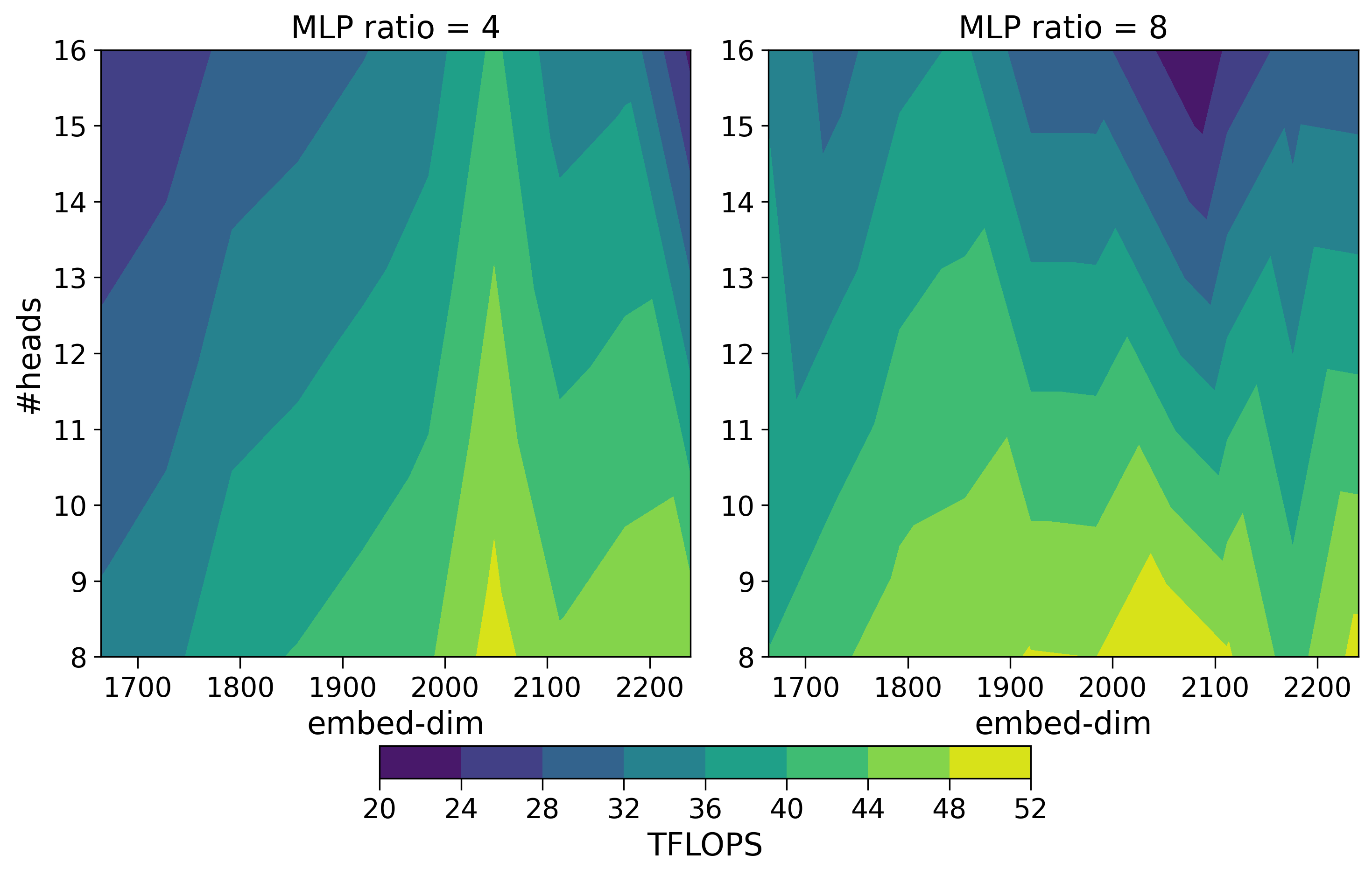}
\vspace{-0.75em}
\caption{Computation performance (TFLOPS) heatmap for the ViT surrogate's architecture on Frontier. }\label{fig:heatmap}
\end{figure}

\paragraph{\bf Compute-efficient architecture} 
As shown in Figure~\ref{fig:heatmap}, the single-node training performance of $256^2$ inputs varies from 20 TFLOPS to 52 TFLOPS, mostly depending on the embedding dimensions, the number of attention heads, and the MLP ratio (i.e., the percentage of MLP parameters of a ViT layer). Typically, higher number of attention heads reduce the performance, and a embedding dimension of 2048 provides the best performance. Increasing the weight of MLP operations will improve the performance overall.     

Based on above heuristics, we design our scaling experiments for three input and model sizes, with detailed architectures listed in Table~\ref{tab:arch}. The number of parameters ranges from 157M to 2.5B. While the number of attention heads is fixed at 8, the embedding dimension increases from 1024 to 2048, to provide more capacity for larger inputs. The number of layers is doubled from each size as well.     
\begin{table}[h]
\scriptsize
\begin{tabular}{c|ccccc|c}
\toprule
input                  & patch & \#layers & \#heads & \#embed dim & \#mlp ratio & \#params \\ \midrule
$64^2$  & 4     & 12       & 8       & 1024        & 4           & 157M     \\
$128^2$ & 4     & 24       & 8       & 2048        & 4           & 1.2B     \\
$256^2$ & 4     & 48       & 8       & 2048        & 4           & 2.5B     \\ \bottomrule
\end{tabular}
\vspace{0.2cm}
\caption{The architecture of the ViT surrogate models.} \label{tab:arch}
\end{table}

\begin{figure}[h!] 
\centering
\includegraphics[width=0.48\textwidth]{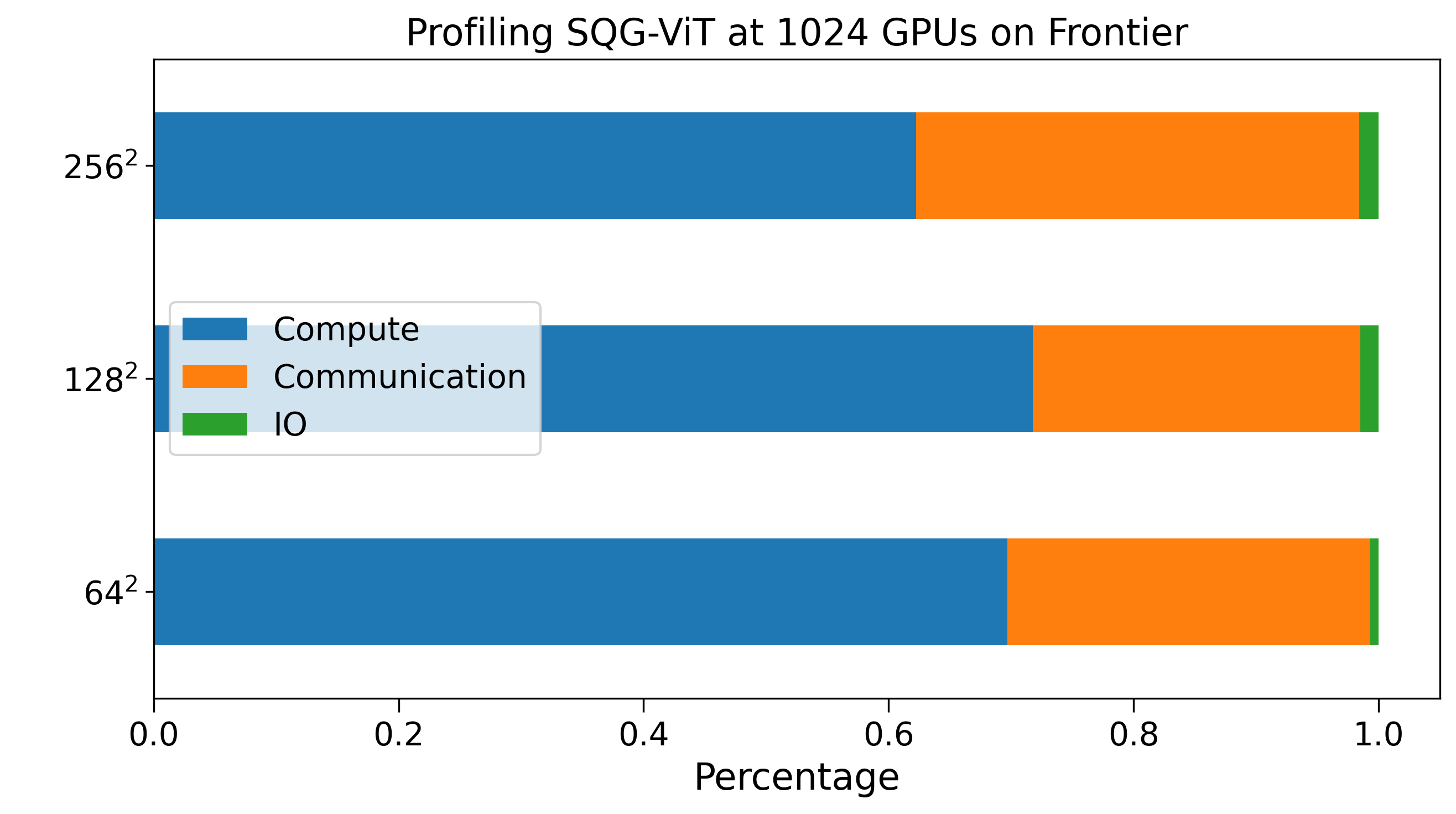}
\caption{The runtime percentage of computation, communication and IO for training theViT surrogate model with input size of $64^2$, $128^2$, and $256^2$, respectively. }\label{fig:profile}
\end{figure}

To study the performance bottleneck, we profile the runtime of the ViT training at 1024 GPUs on Frontier for all three model and input sizes. As shown in Figure~\ref{fig:profile}, the runtime breakdown indicates the training is dominated by computation and communication, with negligible IO, although the IO portion increases slightly from small input ($64^2$) to large input ($256^2$). Specifically, for $64^2$, the computation is less intensive (hence takes longer runtime) compared to larger models due to the $1024$ embedding size, and yet the portion of communication is still larger than that of $128^2$, indicating a slower training performance. On the other hand, for $256^2$, the computation workload is twice of $128^2$, but the communication takes a larger portion because the message volume also doubles. Our results show that ViT training is mostly communication bound at scale, especially for large inputs (i.e., longer sequences).     

\begin{figure}[h!] 
\begin{center}
\includegraphics[width = 0.48\textwidth]{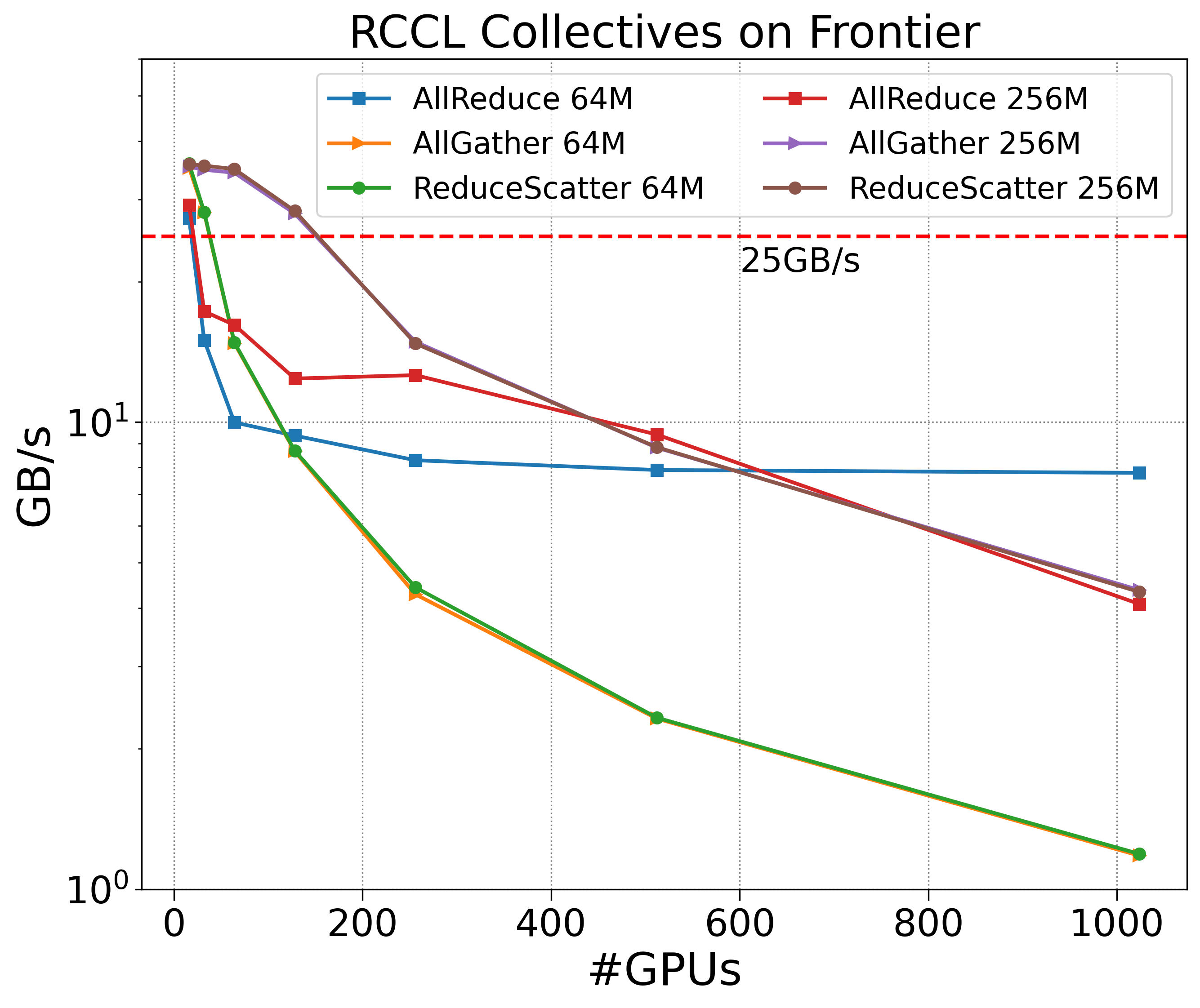}\\
\vspace{0.3cm}
\includegraphics[width = 0.48\textwidth]{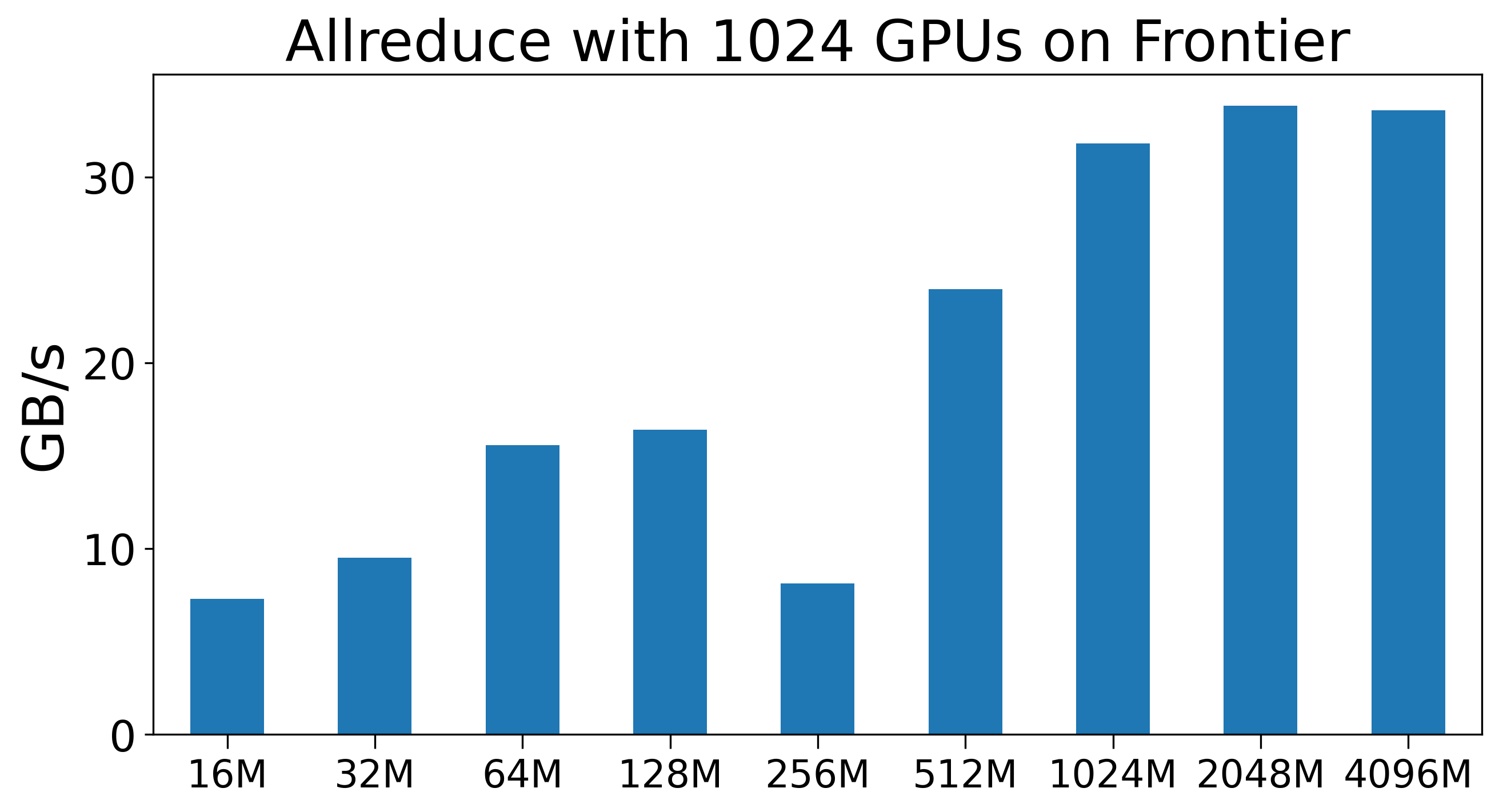}
\end{center}
\vspace{-0.2cm}
\caption{RCCL collectives bandwidth on Frontier. }\label{fig:rccl}
\end{figure}

\paragraph{\bf RCCL Communication}
To establish the communication performance baseline, we measure the RCCL collectives on Frontier. In Figure~\ref{fig:rccl}, we plot the communication bandwidth of \texttt{AllReduce}, \texttt{AllGather}, and \texttt{ReduceScatter} because they are the dominant communication patterns used in data parallelism, including FSDP and ZeRO. For a message size of 64M, the \texttt{AllReduce} significantly outperforms the other two at scale, while for a larger message size, all three schemes perform more or less the same. \texttt{AllGather} and \texttt{ReduceScatter} performs similarly in all cases. Interestingly, while the communication bandwidth improves with message size, there is a sudden performance drop around message size 256MB for \texttt{AllReduce}.    

\begin{figure}[h!] 
\begin{center}
\includegraphics[width = 0.48\textwidth]{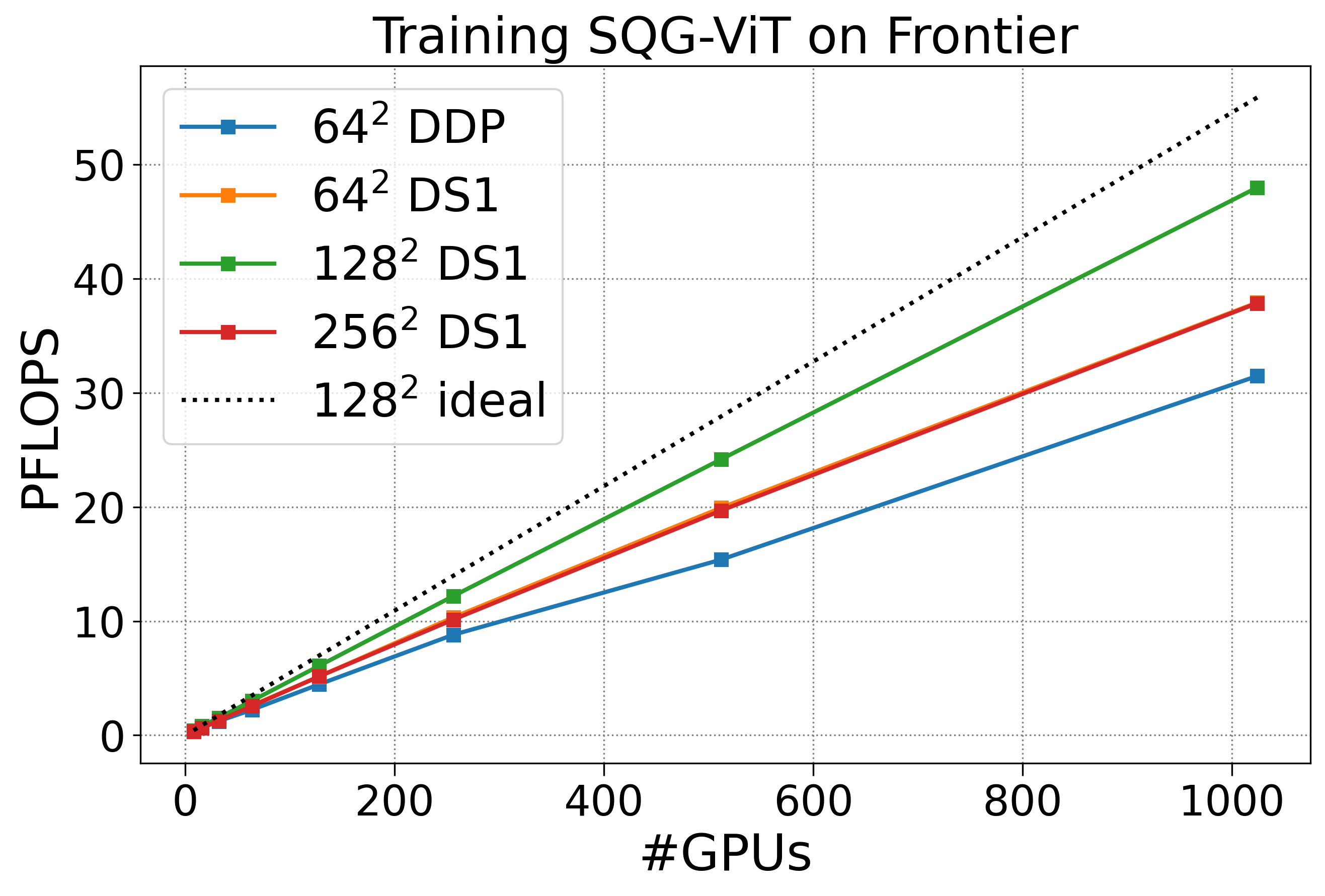}
\includegraphics[width = 0.48\textwidth]{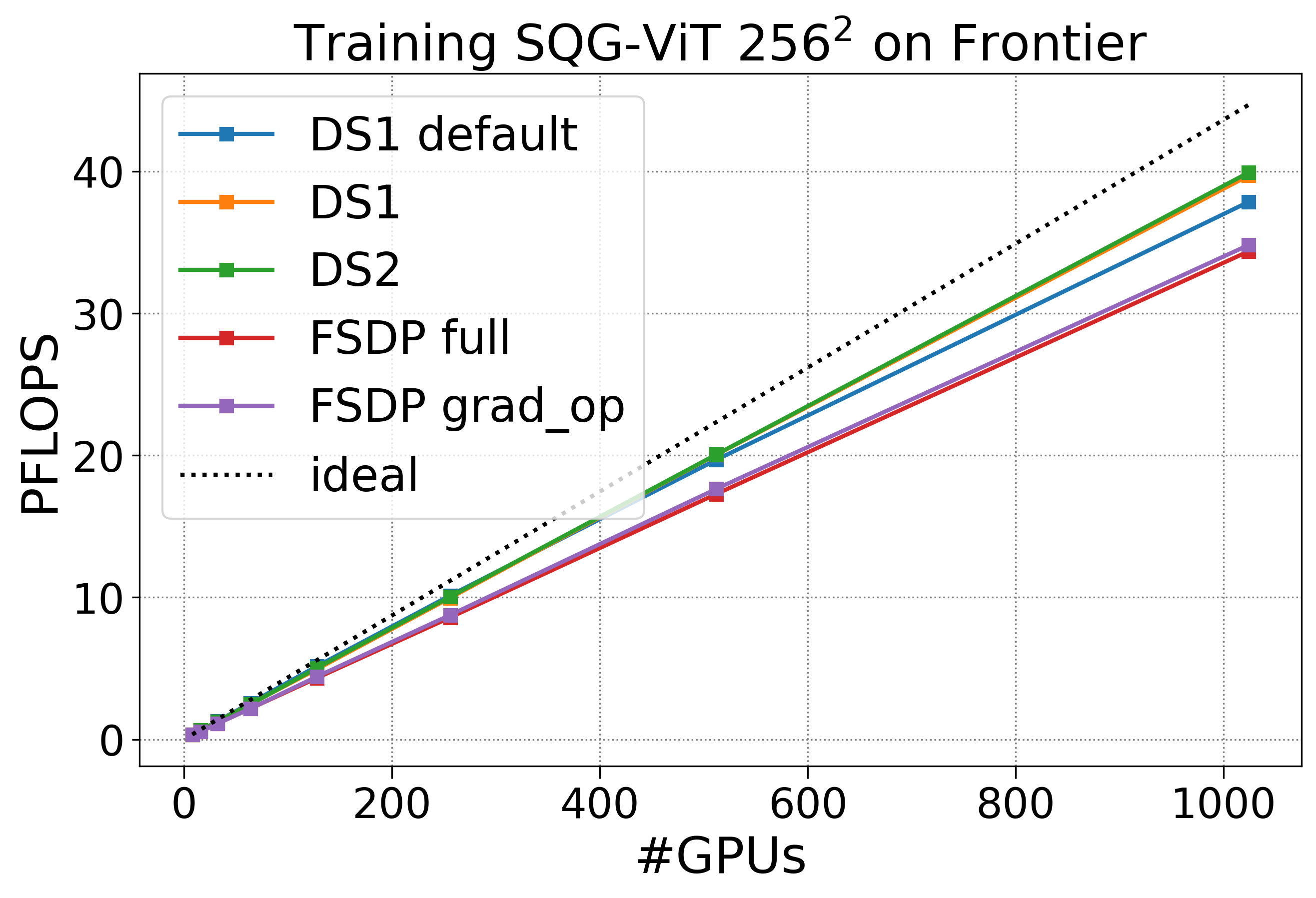}
\end{center}
\vspace{-0.3cm}
\caption{Scaling ViT surrogate up to 1024 GPUs on Frontier with distributed data parallel (DDP), DeepSpeed (DS) stage 1 and 2, and fully sharded data parallel (FSDP) with full and grad\_op strategies. The model size for $64^2$, $128^2$, and $256^2$ input is 157M, 1.2B, and 2.5B, respectively.}\label{fig:scale}
\end{figure}

\paragraph{\bf Scaling on Frontier} With the profiling analysis and baselines established, we are ready to compare different distribution strategies and scale the ViT surrogate up to 1024 GPUs on Frontier. In Figure~\ref{fig:scale}, we first compare the scaling of different model and input sizes. $128^2$ performs the best with a scaling efficiency of 86\%, while $64^2$ and $256^2$ performs comparably. This is consistent with the profiling analysis (see Figure~\ref{fig:profile}), which indicates a trade-off between the computation intensity and communication volume, and $128^2$ input with a 1.2B model size seem to be optimal on Frontier. 

However, for our scientific application, a larger input is desired. To improve the performance of $256^2$, we further study different memory-efficient data-parallel strategies. As shown in Figure~\ref{fig:scale}, the DeepSpeed stage 1 with default setting (message bucket size 200MB) in PyTorch lightning doesn't perform well because the communication bandwidth of \texttt{AllReduce} deteriorates around this message size. On the other hand, a very large message size won't work well either due to less opportunities to overlap communication with computation. We find a message size around 500MB works the best, and resulted scaling efficiency improves to 85\%. Overall, with more optimization knobs, DeeSpeed ZeRO data-parallel outperforms FSDP for training SQG-ViT on Frontier.        

\begin{figure}[h!] 
\begin{center}
\includegraphics[width=0.47\textwidth]{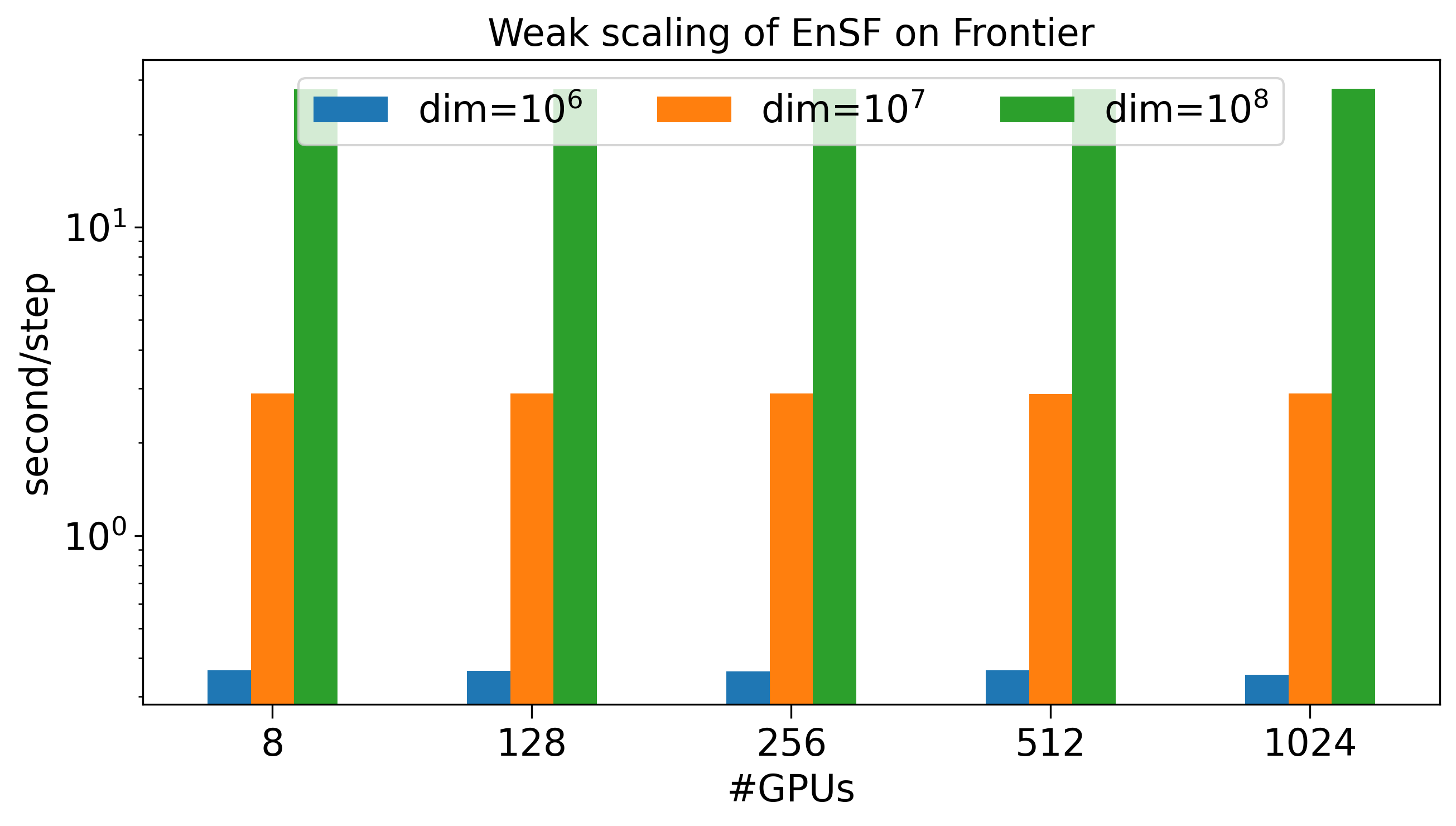}
\end{center}
\vspace{-0.3cm}
\caption{Weak scaling of EnSF on Frontier up to 1024 GPUs for dimension size of $10^6$, $10^7$, and $10^8$, respectively. }\label{fig:ensf}
\end{figure}

\paragraph{\bf EnSF scaling} 
With the training of the forward model optimized, we study the scaling behavior of EnSF on Frontier. The MPI parallelization is along the dimension of the ensemble, so the ranks are straightforwardly parallel and the outputs are MPI reduced in the end. As shown in Figure~\ref{fig:ensf}, EnSF weak scales perfectly up to 1024 GPUs on Frontier. The time per step is about 0.4s for 1M dimension, and 28s for 100M.      

\section{Conclusion}\label{sec:con}
In this study, we introduce a generic sequential data assimilation framework for estimating turbulent dynamics and demonstrate its end-to-end performance on the Frontier supercomputer at OLCF. The system is comprised of a vision transformer (ViT) to emulate the true system evolution and a new ensemble DA method referred to as the ensemble score filter (EnSF). The theoretical basis for EnSF comes from diffusion models which belong to the class of generative AI methods and have the ability to produce highly realistic images and videos. Like other diffusion-based techniques, EnSF leverages the machinery of score functions to represent the complex information in the Bayesian problem. However, the posterior distribution is sampled via a training-free, Monte-Carlo approach which enables us to approximate the corresponding score function directly from the forecast ensemble obtained with the ViT surrogate.

By investigating compute-efficient kernel sizing and comparing various parallelization strategies, we achieve a 85\% strong scaling efficiency and linear weak scaling up to 1024 GPUs, respectively, on the Frontier supercomputer. Our results demonstrate the framework’s exceptional scalability on high-performance computing systems, which is essential for improving the
medium-range forecasts of high-dimensional Earth system applications. As shown in the numerical experiment, e.g., Figure \ref{Linear_shocks_SQG}, physics-based or AI-based weather/climate models cannot predict turbulent dynamics without an efficient DA workflow. The power of the proposed DA framework lies in the fact that it can simultaneously resolve the three main challenges in the geosciences -- nonlinearity/non-Gaussianity, high-dimensionality and scalability on HPC, significantly outperforming SOTA methods like LETKF. We emphasize that the proposed workflow can be combined with any physics-based or AI-based foundation weather models because of using the ViT surrogate. The online training of ViT not only provides an interface with the weather models but also provides the capability of learning from the observation data. Given the outstanding scalability of the our method, the next step is to conducts experiments with more realistic weather models used in operations by working with scientist at the National Oceanic and Atmospheric Administration (NOAA) and European Centre for Medium-Range Weather Forecasts (ECMWF). 

\section*{Acknowledgement}
This work is supported by the U.S. Department of Energy, Office of Science, Office of Advanced Scientific Computing Research, Applied Mathematics program, under the contract ERKJ387, and accomplished at Oak Ridge National Laboratory (ORNL), and under Grant DE-SC0022254. ORNL is operated by UT-Battelle, LLC., for the U.S. Department of Energy under Contract DE-AC05-00OR22725. The first author (FB) would also like to acknowledge the support from U.S. National Science Foundation through project DMS-2142672 and the support from the U.S. Department of Energy, Office of Science, Office of Advanced Scientific Computing Research, Applied Mathematics program under Grant DE-SC0022297.

\end{document}